\documentclass[11pt,letterpaper]{article}

\usepackage[T1]{fontenc}
\usepackage{lmodern}
\usepackage[hmargin=0.82in,top=0.71in,bottom=0.77in]{geometry}
\usepackage{microtype}
\usepackage{amsmath}
\usepackage[table]{xcolor}
\usepackage{graphicx}
\usepackage{booktabs,tabularx}
\usepackage{caption,float}
\usepackage{titlesec,fancyhdr}
\usepackage{hyperref}

\definecolor{ink}{HTML}{182825}
\definecolor{accent}{HTML}{176646}
\definecolor{muted}{HTML}{52645F}
\definecolor{rulegray}{HTML}{C7D2CF}
\definecolor{highlight}{HTML}{EAF4EF}
\hypersetup{
  colorlinks=true, allcolors=accent,
  pdftitle={Tabra Blends: Improving the Accuracy--Latency Trade-off with Complementary Tabular Models},
  pdfauthor={Jing Jia and Guanyang Wang}
}
\titleformat{\section}{\large\sffamily\bfseries\color{ink}}{\thesection}{0.6em}{}
\titlespacing*{\section}{0pt}{1.05em}{0.45em}
\titleformat{\subsection}{\normalsize\sffamily\bfseries\color{ink}}{\thesubsection}{0.6em}{}
\titlespacing*{\subsection}{0pt}{0.8em}{0.35em}
\begin{document}
\color{ink}

{\small\sffamily\bfseries\color{accent}Technical report\hfill November 2026}\par
\vspace{0.18em}
{\color{rulegray}\hrule height 0.6pt}
\vspace{0.6em}
{\fontsize{31}{34}\selectfont\bfseries Tabra Blends}\par
{\large Improving the Accuracy--Latency Trade-off with Complementary Tabular Models}\par
{\small\sffamily
\begin{tabularx}{\linewidth}{@{}XX@{}}
  \textbf{\href{https://jjia131.github.io/}{Jing Jia}} &
    \textbf{\href{https://guanyangwang.github.io/}{Guanyang Wang}} \\
  Department of Computer Science, Rutgers University &
    Department of Statistics, Rutgers University \\
  \href{mailto:jing.jia@rutgers.edu}{jing.jia@rutgers.edu} &
    \href{mailto:guanyang.wang@rutgers.edu}{guanyang.wang@rutgers.edu}
\end{tabularx}}\par

\section*{Abstract}

Tabular foundation models make different trade-offs between prediction accuracy and latency.
We study whether a compact ensemble can deliver higher accuracy in less time. Across 816
TabArena splits, our blend of four-estimator Kumo and one LimiX pipeline reaches 2063.8 Elo,
45.4 points above the top published entry in the same fit. Its normalized fit + predict time
is 15\% lower than Kumo-16's and 98\% lower than full LimiX's. We find that greater prediction
diversity between the two models helps explain the reduction in error.

\section{Introduction}

Tabular foundation models make different trade-offs between prediction accuracy and
latency. Larger ensembles take longer to run, and their accuracy gains can taper
off~\cite{testtime,diversity}. We ask whether a small ensemble can improve the accuracy--latency
frontier.

We combine one LimiX-2-1016 pipeline with varying numbers of Kumo Tabular Large
estimators. The two models rank highest by Elo on the
\href{https://huggingface.co/spaces/TabArena/leaderboard/commit/87bdf9f8c91a21ecfd2a23184aabcf7c75fe612d}{November 20, 2026 TabArena leaderboard}.

\section{Method}

By default, Kumo Tabular Large averages 16 estimators that vary preprocessing and feature
presentation (such as scaling and column order) for the same backbone~\cite{kumo}.
LimiX-2-1016 combines 32 inference pipelines~\cite{limix}.

We use a 50/50 blend of Kumo averaged over \(n \in \{1, 2, 4, 8, 16\}\) estimators and the
first no-retrieval LimiX-2-1016 pipeline for each task type. Both models receive the same
labeled training split and use frozen weights. We average class probabilities for
classification and point predictions for regression.

\section{Evaluation}

\textbf{Benchmark and error.}
We evaluate on 816 official TabArena v0.1 splits across 51 datasets (30 binary, 8
multiclass, and 13 regression)~\cite{tabarena}, using \(1 - \text{ROC AUC}\), log loss, and
RMSE, respectively. Direct head-to-head comparisons average split errors within each dataset
and summarize relative differences across datasets via the geometric mean of error ratios.

\textbf{Elo.}
We fit each local configuration against the same 120 published entries using a
Bradley--Terry model of pairwise split outcomes. Datasets receive equal weight and default
Random Forest is anchored at 1000 Elo. For entries in this pool, Table~\ref{tab:results}
reports ratings from the fit with Kumo-4 + one LimiX pipeline. Elo difference intervals use paired
bootstrap resampling over datasets.

\textbf{Timing.}
For each split, we sum serial fit and predict times across blend constituents. We average
times and row counts by dataset, then scale fit per 1,000 training rows and predict per
1,000 test rows. We report dataset medians for predict and for the sum of the two
normalized stages. Local runs use warmed GH200 timings. Full LimiX, TabPFN-3.5,
TabPFN-3.5-Fast, and TabICLv2 use published timings.

\clearpage
\section{Results}

\begin{figure}[H]
  \centering
  \includegraphics[width=\linewidth]{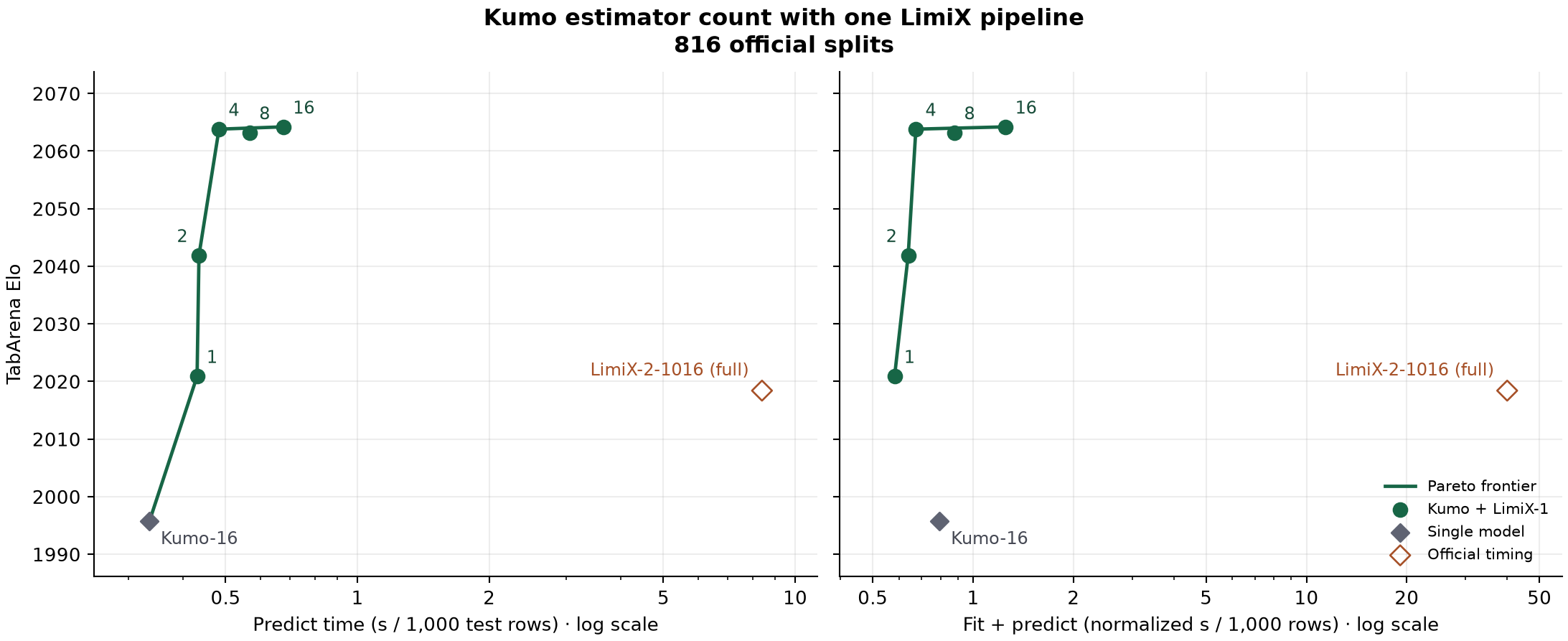}
  \caption{The labels 1, 2, 4, 8 and 16 give the number of Kumo estimators blended with
    one LimiX pipeline. Diamonds mark the original Kumo and LimiX models.}
  \label{fig:pareto}
\end{figure}

\begin{table}[H]
  \caption{\textbf{Accuracy and stage-normalized time on TabArena.}
    Dataset medians in normalized seconds per 1,000 rows. Blends, ablations, and Kumo-16 use
    GH200 timing. Full LimiX, TabPFN-3.5 variants, and TabICLv2 use published TabArena timings.
    The TabPFN-4 preview reports Elo from a separate fit and predict time on RTX PRO 6000.}
  \label{tab:results}
  \small
  \renewcommand{\arraystretch}{1.16}
  \begin{tabularx}{\linewidth}{@{}Xrrr@{}}
    \toprule
    \textbf{Configuration} & \textbf{Elo} & \textbf{Predict} & \textbf{Fit + predict} \\
    \midrule
    \multicolumn{4}{@{}l}{\emph{Single-model ablations}} \\
    Kumo-1 & 1874.2 & 0.075 & 0.148 \\
    Kumo-2 & 1939.1 & 0.085 & 0.190 \\
    Kumo-4 & 1980.6 & 0.125 & 0.285 \\
    Kumo-8 & 1992.6 & 0.223 & 0.490 \\
    LimiX-2-1016 (first pipeline only) & 1922.1 & 0.344 & 0.362 \\
    \addlinespace
    \multicolumn{4}{@{}l}{\emph{Kumo with one LimiX pipeline}} \\
    Kumo-1 + one LimiX pipeline & 2020.9 & 0.430 & 0.581 \\
    Kumo-2 + one LimiX pipeline & 2041.9 & 0.434 & 0.639 \\
    \rowcolor{highlight}
    \textbf{Kumo-4 + one LimiX pipeline} & \textbf{2063.8} & \textbf{0.483} & \textbf{0.673} \\
    Kumo-8~+ one LimiX pipeline & 2063.2 & 0.567 & 0.880 \\
    Kumo-16 + one LimiX pipeline & 2064.2 & 0.678 & 1.252 \\
    \addlinespace
    \multicolumn{4}{@{}l}{\emph{Published references}} \\
    Kumo-16 (GH200 timing) & 1995.7 & 0.335 & 0.793 \\
    LimiX-2-1016 (full) & 2018.4 & 8.398 & 40.022 \\
    \href{https://priorlabs.ai/technical-reports/tabpfn-4-preview}{TabPFN-4} (4 estimators, author-reported) & 1984 & 0.118 & \textemdash \\
    TabPFN-3.5 (default) & 1873.7 & 0.473 & 2.181 \\
    TabPFN-3.5-Fast (default) & 1784.1 & 0.152 & 0.855 \\
    TabICLv2 (default) & 1569.0 & 0.139 & 0.999 \\
    \bottomrule
  \end{tabularx}
\end{table}

\clearpage
\subsection{Accuracy and latency}

Kumo-4 + one LimiX pipeline reaches \textbf{2063.8 Elo} with \textbf{0.673\,s} normalized
fit + predict time. In the same Elo fit, it leads official Kumo-16 by \textbf{68.1 Elo}
(paired 95\% bootstrap interval \(+22.2\) to \(+119.0\)) and full LimiX-2-1016 by
\textbf{45.4 Elo} (\(+14.7\) to \(+75.2\)). The blend uses 15\% less normalized fit + predict
time than Kumo-16 on GH200 and 98\% less than full LimiX's published 40.022\,s. Its
normalized predict time of 0.483\,s falls between locally timed Kumo-16 (0.335\,s) and
full LimiX's published time (8.398\,s).

Elo changes little beyond four Kumo estimators as both timing measures rise. The
eight-estimator blend scores 2063.2 Elo. The 16-estimator blend scores 0.4 Elo higher than
the four-estimator blend, with predict time rising from 0.483\,s to 0.678\,s and fit +
predict from 0.673\,s to 1.252\,s.

At the low-cost end, Kumo-1 + one LimiX pipeline reaches 2020.9 Elo at 0.581\,s normalized
fit + predict time, 27\% below locally timed Kumo-16. Its predict time is 0.430\,s versus
Kumo-16's 0.335\,s.

\subsection{Dataset-level comparisons}

The one-estimator blend records lower error than official Kumo-16 on 33 of 51 datasets
and than the single LimiX pipeline on 37. The four-estimator blend records lower error
than both of its constituents on 32 datasets.

\begin{table}[H]
  \caption{\textbf{Head-to-head results for Kumo-4 + one LimiX pipeline.}
    Each dataset counts once, using its mean error across splits. Error change is the
    geometric mean of per-dataset error ratios, minus one, as a percentage.}
  \label{tab:paired}
  \small
  \renewcommand{\arraystretch}{1.16}
  \begin{tabularx}{\linewidth}{@{}Xrrrr@{}}
    \toprule
    \textbf{Reference} & \textbf{Wins} & \textbf{Losses} & \textbf{Ties} & \textbf{Error change} \\
    \midrule
    Local Kumo-4 & 37 & 14 & 0 & \(-1.43\%\) \\
    First LimiX pipeline & 46 & 5 & 0 & \(-2.41\%\) \\
    Official Kumo-16 & 37 & 14 & 0 & \(-1.06\%\) \\
    Full LimiX-2-1016 & 27 & 24 & 0 & \(-0.10\%\) \\
    \bottomrule
  \end{tabularx}
\end{table}

\section{Prediction complementarity}

We compare two 50/50 blends with Kumo-4. The partners are \textbf{Kumo-16} (1995.2 local
Elo) and the \textbf{first LimiX pipeline} (1922.1 Elo). The LimiX blend achieves lower
error than the Kumo-only control on \textbf{37 of 51 datasets} and lowers geometric-mean
TabArena error by \textbf{1.10\%}. On 14 datasets, adding LimiX improves Kumo-4 even though
the LimiX pipeline has higher standalone error than Kumo-4.

Under squared loss \(S(f)=\mathrm{E}\lVert f-y\rVert^2\) (Brier loss for classification,
MSE for regression), the 50/50 blend's gain over Kumo-4 prediction \(p\) with partner \(q\)
decomposes into two terms~\cite{ambiguity}:
\[
S(p)-S\!\left(\frac{p+q}{2}\right)
=\underbrace{\frac{\mathrm{E}\lVert p-q\rVert^2}{4\mathstrut}}_{\text{disagreement reduction}}
-\underbrace{\frac{S(q)-S(p)}{2\mathstrut}}_{\text{partner penalty}}.
\]
\begin{table}[H]
  \caption{\textbf{Squared-loss gain from adding a 50/50 partner to Kumo-4.}
    Changes are relative to Kumo-4's dataset loss \(S(p)\) and averaged across datasets. A
    negative penalty means the partner has lower standalone loss.}
  \label{tab:mechanism}
  \small
  \renewcommand{\arraystretch}{1.16}
  \begin{tabularx}{\linewidth}{@{}Xrrr@{}}
    \toprule
    \textbf{Partner} & \textbf{Disagreement reduction} & \textbf{Partner penalty} & \textbf{Net gain} \\
    \midrule
    Kumo-16 & \(0.09\%\) & \(-0.15\%\) & \(0.24\%\) \\
    First LimiX pipeline & \(1.24\%\) & \(0.16\%\) & \(1.08\%\) \\
    \bottomrule
  \end{tabularx}
\end{table}

\begin{samepage}
Kumo-16 has lower standalone squared loss and nearly collinear residuals with Kumo-4.
LimiX's median residual cosine distance from Kumo-4 is about 25 times larger. Its larger
disagreement term outweighs the penalty from its higher standalone squared loss, yielding
a \(1.08\%\) net gain. The blend reduces Brier loss by \(0.75\%\) on binary and \(1.45\%\)
on multiclass tasks, and MSE by \(1.61\%\) on regression, relative to Kumo-4.\par
\end{samepage}

\section{Discussion}

\textbf{Direct head-to-head error versus field-wide Elo.}
Against full 32-pipeline LimiX-2-1016, the four-estimator blend is nearly even in direct
dataset-level error (Table~\ref{tab:paired}), winning 27--24 across datasets (16--14 on binary
classification, 5--3 on multiclass classification, and 6--7 on regression) with a \(-0.10\%\)
geometric-mean error change. The 45.4-point Elo lead reflects performance across the field.
Against their 119 shared opponents, the blend scores 95.10\% and full LimiX scores
93.84\% in split comparisons, with half credit for ties and equal weight per dataset.

In separate standalone fits, Kumo-4 scores 1980.6 Elo and the single LimiX pipeline scores
1922.1. Their blend reaches 2063.8 Elo, above all 120 published entries in its fit. This
motivates testing whether training estimators or selecting pipelines for both individual
accuracy and complementary errors improves ensemble accuracy.

\begin{figure}[H]
  \centering
  \includegraphics[width=0.80\linewidth]{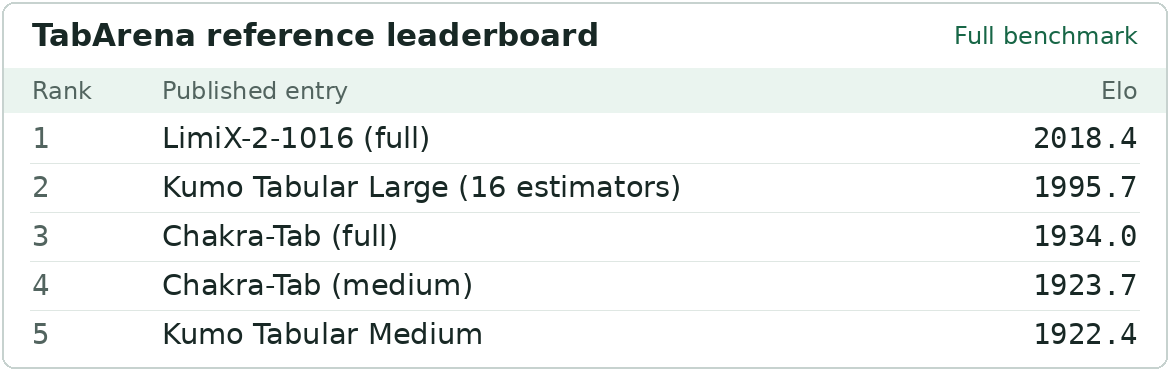}
  \caption{\textbf{Top five published references in our TabArena pool.}
    The scores are from the Elo fit with the four-estimator blend.}
  \label{fig:leaderboard}
\end{figure}

The \href{https://github.com/tabra-ai/tabular-blends}{project repository} contains code and results.

\textbf{Built with StableAI LimiX.}
This is an independent non-commercial research study.

\end{document}